\documentclass[sigconf,natbib=true]{acmart}

\AtBeginDocument{%
  \providecommand\BibTeX{{%
    \normalfont B\kern-0.5em{\scshape i\kern-0.25em b}\kern-0.8em\TeX}}}

\copyrightyear{2022}
\acmYear{2022}
\setcopyright{acmcopyright}\acmConference[SIGIR '22]{Proceedings of the 45th International ACM SIGIR Conference on Research and Development in Information Retrieval}{July 11--15, 2022}{Madrid, Spain}
\acmBooktitle{Proceedings of the 45th International ACM SIGIR Conference on Research and Development in Information Retrieval (SIGIR '22), July 11--15, 2022, Madrid, Spain}
\acmPrice{15.00}
\acmDOI{10.1145/3477495.3531838}
\acmISBN{978-1-4503-8732-3/22/07}

\usepackage{multirow}
\usepackage{array}
\usepackage{bbm}
\usepackage{xcolor}
\usepackage{balance}
\newcommand{\proposed}{\textsf{GraFN}}

\begin{document}
\fancyhead{}
%\settopmatter{printacmref=true, printfolios=false}

%%
%% The "title" command has an optional parameter,
%% allowing the author to define a "short title" to be used in page headers.
\title{\proposed : Semi-Supervised Node Classification on Graph with Few Labels via Non-Parametric Distribution Assignment}

%%
%% The "author" command and its associated commands are used to define
%% the authors and their affiliations.
%% Of note is the shared affiliation of the first two authors, and the
%% "authornote" and "authornotemark" commands
%% used to denote shared contribution to the research.

\author{Junseok Lee}
\orcid{0000-0003-3874-1667}
\affiliation{
    \institution{KAIST ISysE}
    \city{Daejeon}
    \country{Republic of Korea}
}
\email{junseoklee@kaist.ac.kr}
    
\author{Yunhak Oh}
\orcid{0000-0002-9110-3042}
\affiliation{
    \institution{KAIST ISysE}
    \city{Daejeon}
    \country{Republic of Korea}
}
\email{yunhak.oh@kaist.ac.kr}

\author{Yeonjun In}
\orcid{0000-0003-0408-4259}
\affiliation{
    \institution{KAIST ISysE}
    \city{Daejeon}
    \country{Republic of Korea}
}
\email{yeonjun.in@kaist.ac.kr}

\author{Namkyeong Lee}
\orcid{0000-0003-3995-1148}
\affiliation{
    \institution{KAIST ISysE}
    \city{Daejeon}
    \country{Republic of Korea}
}
\email{namkyeong96@kaist.ac.kr}

\author{Dongmin Hyun}
\orcid{0000-0001-7757-3227}
\affiliation{
    \institution{POSTECH PIAI}
    \city{Pohang}
    \country{Republic of Korea}
}
\email{dm.hyun@postech.ac.kr}

\author{Chanyoung Park}
\orcid{0000-0002-5957-5816}
\affiliation{
    \institution{KAIST ISysE \& AI}
    \city{Daejeon}
    \country{Republic of Korea}
}
\email{cy.park@kaist.ac.kr}
\authornote{
Corresponding author.
}

\renewcommand{\shortauthors}{Lee, et al.}

%%
%% By default, the full list of authors will be used in the page
%% headers. Often, this list is too long, and will overlap
%% other information printed in the page headers. This command allows
%% the author to define a more concise list
%% of authors' names for this purpose.
% \renewcommand{\shortauthors}{Trovato and Tobin, et al.}

%%
%% The abstract is a short summary of the work to be presented in the
%% article.
\begin{abstract}
Despite the success of Graph Neural Networks (GNNs) on various applications, GNNs encounter significant performance degradation when the amount of supervision signals, i.e., number of labeled nodes, is limited, which is expected as GNNs are trained solely based on the supervision obtained from the labeled nodes. On the other hand, recent self-supervised learning paradigm aims to train GNNs by solving pretext tasks that do not require any labeled nodes, and it has shown to even outperform GNNs trained with few labeled nodes. However, a major drawback of self-supervised methods is that they fall short of learning class discriminative node representations since no labeled information is utilized during training. 
To this end, we propose a novel semi-supervised method for graphs,~\proposed\footnote{The source code for ~\proposed~ is available at \href{https://github.com/Junseok0207/GraFN}{https://github.com/Junseok0207/GraFN}} that leverages few labeled nodes to ensure nodes that belong to the same class to be grouped together, thereby achieving the best of both worlds of semi-supervised and self-supervised methods. Specifically,~\proposed~randomly samples support nodes from labeled nodes and anchor nodes from the entire graph. Then, it  minimizes the difference between two predicted class distributions that are non-parametrically assigned by anchor-supports similarity from two differently augmented graphs. 
We experimentally show that~\proposed~surpasses both the semi-supervised and self-supervised methods in terms of node classification on real-world graphs.
%The source code for ~\proposed~ is available at \href{https://github.com/Junseok0207/GraFN}{https://github.com/Junseok0207/GraFN}.
\vspace{-1ex}
\end{abstract}

%%
%% The code below is generated by the tool at http://dl.acm.org/ccs.cfm.
%% Please copy and paste the code instead of the example below.
%%


%% Keywords. The author(s) should pick words that accurately describe
%% the work being presented. Separate the keywords with commas.
\vspace{-1ex}

%%
%% This command processes the author and affiliation and title
%% information and builds the first part of the formatted document.
\maketitle
\begin{figure}[h]
  \includegraphics[width=0.99\linewidth]{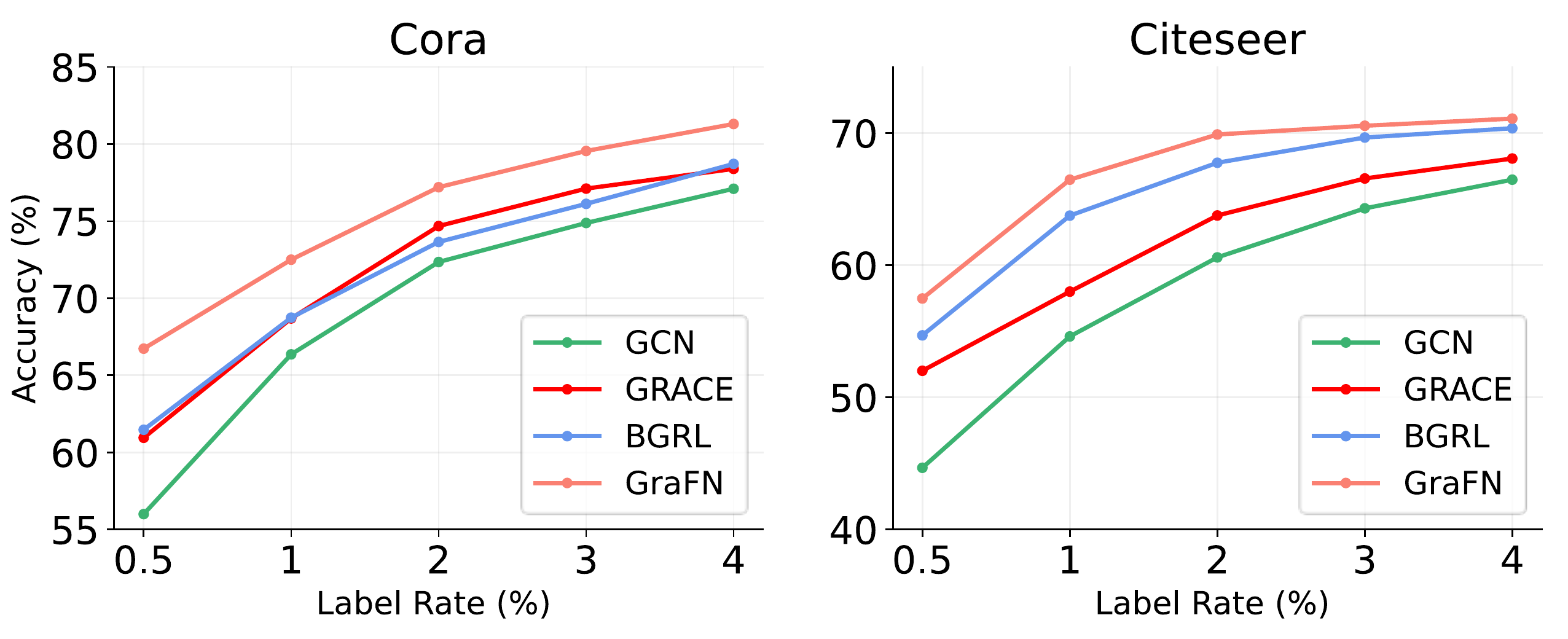}
  \vspace{-1ex}
  \caption{Performance of GCN, GRACE, BGRL and~\proposed~on Cora and Citeseer datasets over various labeled node rates.}
  \vspace{-3ex}
  \label{fig:teaser}
\end{figure}

\section{Introduction}

% Over the past few years, leveraging graph structured data has received a surge of interest with its inherent property that captures the relationship between data instances, i.e., nodes. 
Recently, Graph Neural Networks (GNNs) are widely applied in IR applications ranging from recommender system~\cite{he2020lightgcn, liu2021interest, wu2021self, jin2020multi, wang2019neural}, question answering~\cite{hu2020residual,zhang2020answer} to web search~\cite{mao2020item, lin2021graph}.
% a plethora of research has been conducted to handle graphs, and applied to various downstream tasks such as node classification \textcolor{red}{CITE} and link prediction \textcolor{red}{CITE}. 
% Recently, a plethora of research has been conducted to handle graphs, and applied to various downstream tasks such as node classification \textcolor{red}{CITE} and link prediction \textcolor{red}{CITE}. 
Specifically, graph convolution-based methods~\cite{zhao2021wgcn,xu2020label} incorporate rich attributes of nodes along with the structural information of graphs by recursively aggregating the neighborhood information. 
% With sufficiently annotated labels, GNNs have shown to be effective for node classification.
Despite the success, the performance of GNNs on node classification significantly degrades when only few labeled nodes are given as GNNs tend to overfit to the few labeled nodes. To make the matter worse, GNNs suffer from ineffective propagation of supervisory signals due to the well-known over-smoothing issue~\cite{deeper}, which makes GNNs not being able to fully benefit from the given labeled nodes. Fig.~\ref{fig:teaser} demonstrates a severe performance degradation of GCN as the rate of the labeled nodes decreases. 

\begin{figure*}[t]
% \begin{figure}[!htbp]
  \includegraphics[width=0.7\linewidth]{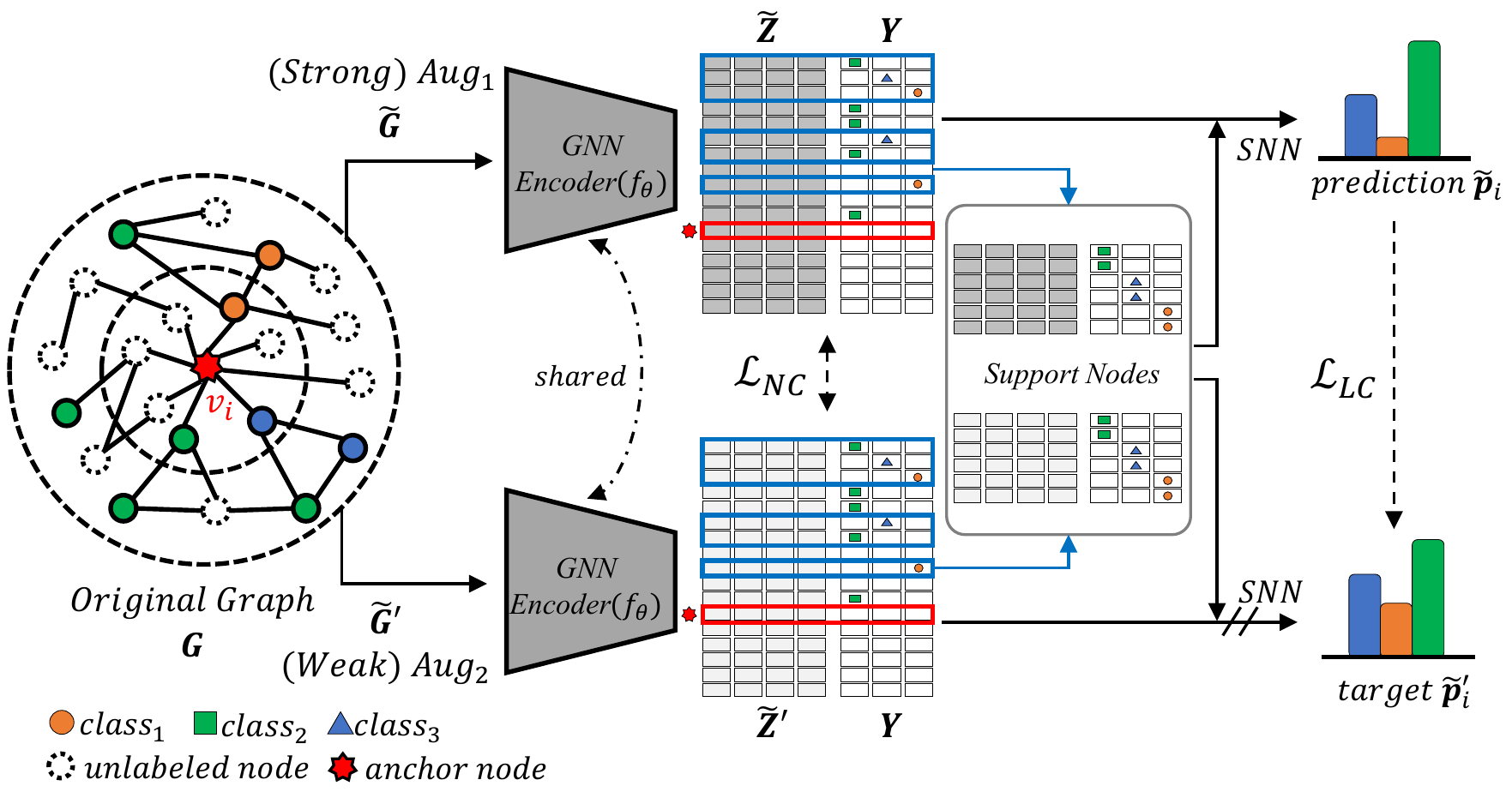}
%   \vspace{-1ex}
  \caption{
  The overall model architecture of~\proposed. Given a graph $G$, we generate two differently augmented views $\tilde{G}$ and $\tilde{G}^{'}$ both of which are fed into a shared encoder $f_{\theta}$ to obtain node-level representation $\tilde{\mathbf{Z}}$ and $\tilde{\mathbf{Z}}^{'}$, respectively. Then,~\proposed~not only minimizes the difference between these two representations obtained from differently augmented graphs, i.e. node-wise consistency ($\mathcal{L}_\text{NC}$), but also minimizes the difference between the predicted class distributions computed in a non-parametrically manner by using the similarity between the anchor node and support nodes, which are randomly sampled from labeled nodes, i.e., label-guided consistency ($\mathcal{L}_\text{LC}$).
%   enforcing them to be consistently close with a certain class of labeled nodes, i.e. Label-guided Consistency. To do this, we randomly sample partial labeled nodes(support nodes) and compute predicted class distribution using anchor-support similarity. 
  }
    \label{fig:overall}
\end{figure*}

To train GNNs given limited labeled information, recent studies mainly focus on leveraging pseudo-labeling techniques. Specifically, co-training~\cite{deeper}~uses Parwalks~\cite{parwalks}~to provide confident pseudo-labels to help train GNNs, and self-training~\cite{deeper} expands the label set by obtaining pseudo-labels provided by GNNs trained in advance. Moreover, M3S~\cite{M3S} leverages a clustering technique to filter out pseudo-labels that do not align with the clustering assignments for improving the pseudo-labeling accuracy. 
However, pseudo-labeling-based methods suffer from an inherent limitation originated from the incorrect pseudo-labels, which eventually incur confirmation bias~\cite{confirmation}. To alleviate this issue, it is crucial to fully benefit from the given label information.

On the other hand, self-supervised methods for graphs~\cite{GRACE,GCA,BGRL} learn node representations without any requirements of labeled nodes. In particular, based on graph augmentations, contrastive learning-based methods pull positive pairs of nodes together while pushing negative ones apart, whereas consistency regularization-based methods impose GNNs to consistently output the same node representations over various perturbations of the given graph. Although these methods have achieved the state-of-the-art results in node classification even outperforming the supervised counterparts, they fall short of learning class discriminative node representations since no labeled information is utilized during training. As shown in Fig.~\ref{fig:teaser}, although recent self-supervised methods for graphs, i.e., BGRL~\cite{BGRL} and GRACE~\cite{GRACE}, 
outperform GCN~\cite{GCN} over various rates of labeled nodes, the performance degradation is still severe as the rate decreases.

To this end, we propose a simple yet effective semi-supervised method for graphs that fully leverages a small amount of labeled nodes to learn class discriminative node representations, called \proposed. The main idea is to consistently impose the representations of nodes that belong to the same class to be grouped together on differently augmented graphs. Specifically, we randomly sample support nodes from the labeled nodes and anchor nodes from the entire graph, and non-parametrically compute two predicted class distributions from two augmented graphs based on the anchor-supports similarity. By minimizing the difference between the two class distributions,~\proposed~not only learns augmentation invariant parameters, but also enforces the representations of nodes that belong to the same class to be grouped together. As shown in Fig.~\ref{fig:teaser},~\proposed~consistently outperforms both the semi-supervised and self-supervised baselines over various rates of labeled nodes, especially outperforming when the number of labeled nodes is smaller, which demonstrates the robustness of~\proposed.

% \section{PRELIMINARIES}
\smallskip
\noindent\textbf{Notations}. Let $G=(V, E)$ denote a graph, where $V$ is the set of $|V|=N$ nodes and $E$ is the set of edges between the nodes. The adjacency matrix is defined by $\mathbf{A} \in \mathbb{R}^{N \times N}$ with each element $\mathbf{A}_{ij}=1$ indicating the existence of an edge between nodes $v_i$ and $v_j$, otherwise $\mathbf{A}_{ij}=0$.
The node attributes are denoted by $\mathbf{X} \in \mathbb{R}^{N \times F}$, where $F$ is the number of features of each node. Additionally, the label matrix is denoted by $\mathbf{Y}\in\mathbb{R}^{N\times C}$, where $C$ is the number of classes, and each row, i.e., $\mathbf{Y}_i\in\mathbb{R}^C$, is the one-hot label vector for node $v_i$. 
We denote $V_L$ and $V_U$ as the set of labeled and unlabeled nodes, respectively.
% We denote $V_L$ and $V_U$ as the set of labeled and unlabeled nodes, respectively, where $|V_{L}|=N_{L}$, $|V_{U}|=N_{U}$, $V_L\cup V_U = V$, and $V_L\cap V_U = \emptyset$.
% Our goal is to accurately predict the labels of nodes that belong to $V_U$ given few labeled nodes, i.e., $N_{L} \ll N_{U}$.
Our goal is to accurately predict the labels of nodes that belong to $V_U$ given few labeled nodes, i.e., $|V_{L}| \ll |V_{U}|$.
% We denote $|V_{L}|=N_{L}$ and $|V_{U}|=N_{U}$ as the set of labeled and unlabeled nodes, respectively. We assume few labeled node are given, $N_{L} \ll N_{U}$. Our goal is to accurately predict the labels of nodes that belong to $V_U$.
\vspace{-1ex}

\begin{table*}[t]
% \small
\centering
\caption{Test Accuracy on semi-supervised node classification.}
\vspace{-2ex}
\renewcommand{\arraystretch}{0.95}
\begin{tabular}{l|ccc|ccc|ccc|ccc|ccc}
    \noalign{\smallskip}\noalign{\smallskip}
    \toprule
    %\hline
    %\multirow{2}{*}{Methods}& \multicolumn{3}{c}{Cora} & \multicolumn{3}{c}{Citeseer} &  \multicolumn{3}{c}{Pubmed} & \multicolumn{3}{c}{Am. Comp} & \multicolumn{3}{c}{Am. Photos} \\
    {Methods} & \multicolumn{3}{c}{Cora} & \multicolumn{3}{c}{Citeseer} &  \multicolumn{3}{c}{Pubmed} & \multicolumn{3}{c}{Am. Comp} & \multicolumn{3}{c}{Am. Photos} \\
    \hline
    %\cline{2-16}
    %\cline{2-4} \cline{6-8} \cline{10-12} \cline{14-16} \cline{18-20} \cline{22-24}
    {Label Rate} & 0.5\% & 1\% & 2\% & 0.5\% & 1\% & 2\% & 0.03\% & 0.06\% & 0.1\% & 0.15\% & 0.2\% & 0.25\% & 0.15\% & 0.2\% & 0.25\% \\
    \midrule
    %\hline
    MLP & 31.24 & 37.74 & 44.53 & 32.07 & 43.07 & 46.11 & 52.50 & 55.80 & 61.22 & 40.30 & 42.22 & 49.98 & 29.76 & 31.64 & 38.55 \\
    LP & 50.77 & 58.28 & 64.43 & 31.15 & 37.95 & 41.71 & 50.93 & 55.83 & 62.14 & 60.46 & 65.90 & 68.79 & 63.67 & 66.38 & 70.40 \\
    GCN & 56.00 & 66.36 & 72.35 & 44.67 & 54.61 & 60.59 & 59.28 & 64.00 & 73.74 & 62.71 & 66.81 & 71.75 & 66.70 & 70.72 & 75.74\\  
    \hline
    GAT & 58.57 & 67.75 & 72.74 & 48.70 & 58.73 & 62.71 & 63.15 & 64.11 & 73.19 & 66.17 & 70.18 & 72.82 & 73.29 & 74.46 & 80.12 \\
    SGC & 49.19 & 63.60 & 69.56 & 44.02 & 55.89 & 63.61 & 58.58 & 62.50 & 71.90 & 59.69 & 64.24 & 68.29 & 55.96 & 61.64 & 69.69 \\
    APPNP & 62.02 & 71.45 & 76.89 & 41.79 & 54.70 & 62.86 & 63.15 & 64.11 & 73.19 & 68.53 & 72.47 & 74.27 & 75.54 & 78.49 & 82.75 \\
    GRAND & 54.51 & 70.92 & 74.90 & 46.76 & 58.40 & 65.31 & 55.87 & 61.25 & 72.42 & 68.00 & 72.71 & 75.77 & 73.80 & 75.83 & 82.33 \\
    \hline
    GLP & 56.94 & 68.28 & 72.97 & 41.53 & 54.84 & 63.08 & 56.70 & 60.83 & 73.46 & 62.97 & 68.56 & 70.70 & 63.18 & 67.96 & 75.19 \\
    IGCN & 58.81 & 70.10 & 74.34 & 43.28 & 57.00 & 64.62 & 57.50 & 62.06 & 73.13 & 65.48 & 70.05 & 71.03 & 71.27 & 73.28 & 77.93 \\
    CGPN & 64.21 & 70.54 & 72.97 & 53.90 & 63.70 & 65.15 & 64.55 & 67.58 & 71.42 & 65.37 & 67.98 & 70.77 & 74.14 & 76.89 & 81.57 \\
    \hline
    GRACE & 60.95 & 68.69 & 74.68 & 52.01 & 58.00 & 63.76 & 64.86 & 68.35 & \textbf{75.92} & 65.25 & 67.79 & 71.79 & 70.19 & 71.89 & 77.32 \\
    BGRL & 61.74 & 68.74 & 73.65 & 54.69 & 63.75 & 67.75 & 65.77 & \textbf{68.86} & 75.91 & 68.80 & 73.04 & 75.11 & 74.27 & 78.25 & 83.12 \\
    \hline
    Co-training & 62.75 & 68.72 & 74.05 & 43.76 & 54.75 & 61.13 & 63.01 & 68.15 & 74.24 & 67.06 & 71.62 & 71.34 & 72.85 & 74.65 & 79.92 \\
    Self-training & 57.28 & 70.73 & 75.40 & 46.26 & 60.36 & 66.47 & 57.34 & 65.13 & 72.86 & 61.32 & 65.95 & 68.66 & 61.92 & 65.24 & 71.34 \\
    M3S & 64.46 & \textbf{72.93} & 76.41 & 55.07 & 65.74 & 67.64 & 61.53 & 64.60 & 73.18 & 61.51 & 66.30 & 68.10 & 63.93 & 67.62 & 73.39 \\
    \hline
    \hline
    \textbf{\proposed} & \textbf{66.73} & 72.50 & \textbf{77.20} & \textbf{57.48} & \textbf{66.47} & \textbf{69.89} & \textbf{65.91} & 68.41 & 75.74 & \textbf{71.73} & \textbf{74.26} & \textbf{77.37} & \textbf{79.25} & \textbf{80.87} & \textbf{85.36} \\
    \bottomrule
\end{tabular}
\vspace{-1ex}
\label{tab:main_table}
\end{table*}

%\section{OVERALL METHODOLOGY}
%\section{Proposed Method:~\proposed}
\section{Proposed Method : \textsf{G\lowercase{ra}FN}}
\label{sec:method}
\smallskip
\noindent\textbf{1) Graph Augmentations and Encoding.}
Given a graph, we first generate two graph views by applying stochastic graph augmentations, which randomly mask node features and drop partial edges. Two differently augmented views are denoted by 
$\tilde{G}=(\tilde{\mathbf{A}}, 
\tilde{\mathbf{X}})$ and $\tilde{G}^{'}=(\tilde{\mathbf{A}}^{'}, \tilde{\mathbf{X}}^{'})$.
% called anchor view and positive view respectively. 
Each augmented view is fed into a \textit{shared} GNN encoder, $f_{\theta}: {\mathbb{R}^{N \times N}} \times \mathbb{R}^{N \times F} \rightarrow \mathbb{R}^{N \times D}$,  to obtain low dimensional node-level representations
$f_\theta(\tilde{\mathbf{A}}, \tilde{\mathbf{X}})=\tilde{\mathbf{Z}}\in\mathbb{R}^{N\times D}$,  and $f_\theta(\tilde{\mathbf{A}}^{'}, \tilde{\mathbf{X}}^{'})=\tilde{\mathbf{Z}}^{'}\in\mathbb{R}^{N\times D}$. Note that we adopt GCN as the backbone of the GNN encoder.
% $f_\theta(\tilde{\mathbf{A}}, \tilde{\mathbf{X}})=\tilde{\mathbf{Z}} = \{{\mathbf{\tilde{z}}_1}, {\mathbf{\tilde{z}}_2}, ..., {\mathbf{\tilde{z}}_n} \}\in\mathbb{R}^{N\times D}$,  and $f_\theta(\tilde{\mathbf{A}}^{'}, \tilde{\mathbf{X}}^{'})=\tilde{\mathbf{Z}}^{'} = \{{\mathbf{\tilde{z}}_1}^{'}, {\mathbf{\tilde{z}}_2}^{'}, ..., {\mathbf{\tilde{z}}_n}^{'} \}\in\mathbb{R}^{N\times D}$. 
% Then, we instance wisely minimize difference between differently augmented graph to learn augmentation invariant representation. 

\smallskip
\noindent{\textbf{2) Node-wise Consistency Regularization.}}
Then, to learn augmentation invariant node representations, we minimize the difference, i.e., cosine distance, between the representations obtained from the two differently augmented graphs in a node-wise manner:
\begin{equation}
\label{eqn:nc}
% \small
\mathcal{L}_{\text{NC}} = -\frac{1}{N}\sum_{i=1}^{N}\frac{\mathbf{\tilde{Z}}_{i}\cdot\mathbf{\tilde{Z}}^{'}_{i}}{\| \mathbf{\tilde{Z}}_{i}\| \|\mathbf{\tilde{Z}}^{'}_{i} \|} 
\vspace{-1ex}
\end{equation}
Note that the above loss can be considered as a simplified version of the self-supervised loss proposed in BGRL.
The major difference is that BGRL involves two separate encoders, where one encoder is updated by minimizing the distance between the node representations obtained from the two views, while the other one is updated by the exponential moving average of the parameters of the other encoder to prevent the collapsing of node representations to trivial solutions.
On the other hand,~\proposed~trains only one \textit{shared} encoder, and we find that the supervisory signals incorporated in the next step help avoid the collapse of representations.
% , which minimizes the distance between the node representations obtained from the two views with two separate encoders.
% To prevent the collapsing of node representations to trivial solutions,

% The major difference is that BGRL uses two encoders, and  where one encoder is updated by minimizing the distance between the node representations from the two views while the other one is updated by the exponential moving average of the parameters of the other encoder, whereas~\proposed~relies on a single encoder.
% However,~\proposed~does not use additional teacher network or prediction layer such as BGRL, because 
% We find that additional supervisory signals in the next stage helps avoid the collapse of representations. 

\smallskip
\noindent{\textbf{3) Label-guided Consistency Regularization.}}
% Although the above self-supervised loss has shown to be effective, it can incur relatively not discriminative node representation because it cannot leverage the node label information. To achieve best of both worlds of self-supervised and semi-supervised learning, we infuse label information by enforcing consistency between predicted class distribution computed by non-parametric classifier $\pi_{d}$ using Soft Nearest Neighbours(SNN) strategy~\cite{snn}. The main idea is that preserving predicted class distribution obtained by similarity with labeled nodes on differently augmented graphs can help node representation belongs to same class to be grouped together. Specifically, we randomly sample the same number of labeled nodes per each class to construct the support set at each iteration $S$ and we denote $Z_S=\{z_{S_{1}}, z_{S_{2}}, ... , z_{S_{b\times C}}\}$ nodes in support set, where $b$ is number of samples per each class. To enjoy large support set $S$, we fix $b$ to the number of labeled nodes belongs to minority class. Then we adopt Soft Nearest Neighbours strategy to non-parametrically assign class distribution using anchor-support similarity. The similarity classifier $\pi_{d}$ is formally expressed as 
Although the above self-supervised loss has been shown to be effective, the learned node representations are not class discriminative because the node label information is not involved in the training process. 
We argue that unlabeled nodes can be grouped together according to their classes by enforcing them to be consistently close with a certain class of labeled nodes on differently augmented graphs. 
% For example, when we know the interests of James and Max, when Arnold has same interest with James and Max, if Arnold is distant from James in a certain augmentation, 
% \textcolor{red}{(TODO: Give a toy example explaining what this means.)}\textcolor{blue}{We argue that unlabeled nodes can be tightly grouped together by enforcing unlabeled nodes to consistenly close with certain class of labeled nodes on differently augmented graph. For example, when we know the interests of James and Max, when Arnold has same interest with James and Max, if Arnold is distant from James in a certain augmentation, }
Hence, we compute the similarity between few labeled nodes and all the nodes in the two augmented graphs, and maximize the consistency between the two similarity-based class assignments, expecting that this would help nodes that belong to the same class to be grouped together.
% maximize the consistency between the predicted class distributions computed based on the similarity of nodes with labeled nodes on differently augmented graphs.  can help nodes that belong to same class to be grouped together. 
% Hence, maximizing the consistency between the predicted class distributions computed based on the similarity of nodes with labeled nodes on differently augmented graphs can help nodes that belong to same class to be grouped together. 
More precisely, we first randomly sample $b$ labeled nodes per class to construct the support set $\mathcal{S}$, and let $\mathbf{Z}^\mathcal{S}\in\mathbb{R}^{(b\times C)\times D}$ denote ($b\times C$) support node representations\footnote{To make the best use of labeled nodes in the support set $S$, we fix $b$ to the number of labeled nodes that belong to the class with the fewest nodes, i.e., minority class.}. 
% and we denote $\mathbf{Z}_S=\{\mathbf{z}_{S_{1}}, \mathbf{z}_{S_{2}}, ... , \mathbf{z}_{S_{b\times C}}\}\in\mathbb{R}^{(b\times C)\times D}$ as the representations of nodes in the support set where $b$ is number of sampled nodes per class. 
Then, for each anchor node $v_i\in V$, we compute the similarity distribution, i.e., predicted class distribution, $\mathbf{p}_i \in \mathbb{R}^{C}$ using anchor-supports similarity in a non-parametric manner by applying Soft Nearest Neighbors(SNN) strategy~\cite{snn, paws} as follows:
% Then, we assign one-hot distribution for labeled nodes, otherwise, non-parametrically assign predicted class distribution $\mathbf{p}_i \in \mathbb{R}^{C}$ using anchor-support similarity for unlabeled node.
% To do this, we apply Soft Nearest Neighbours(SNN) strategy~\cite{snn}
% \pi_d(\mathbf{z}_{i}, \mathbf{Z}_S)
\begin{equation}
% \small
\label{eqn:snn}
\mathbf{p}_i =\sum_{(\mathbf{Z}^\mathcal{S}_{j}, \mathbf{Y}^\mathcal{S}_{j}) \in \mathbf{Z}^\mathcal{S}}{\frac{\exp{(\textsf{sim}(\mathbf{Z}_{i}, \mathbf{Z}^\mathcal{S}_{j})/\tau)}}{\sum_{\mathbf{Z}^\mathcal{S}_{k}\in\mathbf{Z}^\mathcal{S}} \exp{({\textsf{sim}(\mathbf{Z}_{i}, \mathbf{Z}^\mathcal{S}_{k})}}/\tau)}}\cdot \mathbf{Y}^\mathcal{S}_j
\end{equation}
where $\textsf{sim}(\cdot, \cdot)$ computes the cosine similarity between two vectors, and $\tau$ is a temperature hyperparameter. 
$\mathbf{p}_i$ can be considered as the soft pseudo-label vector for node $v_i$ since it is derived based on the labeled nodes in $\mathcal{S}$.
Having defined the predicted class distribution as in Eqn.~\ref{eqn:snn}, we compute the predicted class distributions for each node $v_i\in V$, i.e., $\mathbf{\tilde{p}}_i$ and $\mathbf{\tilde{p}^{'}}_i$ each of which is obtained from $\tilde{G}$ and $\tilde{G}^{'}$, respectively, where the former is considered as the prediction distribution, and the latter is considered as the target distribution.
Then,
% we obtain the predicted class distributions for node $v_i$ computed based on $\tilde{G}$ and $\tilde{G}^{'}$ are $\mathbf{\tilde{p}}_i$ and $\mathbf{\tilde{p}^{'}}_i$, respectively.
% By doing so, the label information can be effectively propagated to distant nodes. 
% It is important to note that we enforce the model to output a low-entropy prediction of class distribution $\mathbf{p}_i$ by adopting a sharpening operation as follows:
% \begin{equation}
% [Sharpen(\mathbf{p}_{i})]_c = \frac{{[\mathbf{p}_{i}]_{c}}^{\frac{1}{T}}}{\sum_{j=1}^{C}{[\mathbf{p}_{i}]_{j}}^{\frac{1}{T}}},     c = 1,...,C
% \end{equation}
% where $T$ is a temperature hyperparameter. As $T$ decreases, the entropy of $\mathbf{p}_i$ becomes low, which implies that the prediction is confident.
% for sharp predicted class distribution. 
% given the two predicted class distributions, i.e., $\tilde{\mathbf{p}}_{i}$ and $\tilde{\mathbf{p}}_{i}^{'}$,
% each of which is computed from $\tilde{G}$ and $\tilde{G}^{'}$, respectively, 
we minimize the cross-entropy between them:
\begin{equation}
% \small
\label{eqn:lc_1}
\frac{1}{|V_U|}\sum_{v_i\in V_U}H(\tilde{\mathbf{p}}_{i}^{'},\tilde{\mathbf{p}}_{i}) + \frac{1}{|V_L|}\sum_{v_i\in V_L}H(\mathbf{Y}_i,\tilde{\mathbf{p}}_{i})
% \frac{1}{|V|}\sum_{v_i\in V}H(\tilde{\mathbf{p}}_{i},\tilde{\mathbf{p}}_{i}^{'})
\end{equation}
where $H(\mathbf{y},\mathbf{\hat{y}})$ is the cross-entropy between the target $\mathbf{y}$ and the prediction $\mathbf{\hat{y}}$.
Note that for each labeled node, instead of computing the predicted class distribution as in Eqn.~\ref{eqn:snn}, we simply assign its one-hot label vector, i.e., $\mathbf{\tilde{p}^{'}}_i=\mathbf{Y}_i, \,\forall v_i\in V_L$, to fully leverage the label information.
% \textcolor{blue}{We argue that unlabeled nodes can have high quality target distribution $\mathbf{\tilde{p}^{'}}_i$ which sharpens on corresponding class by combining with node-wise consistency regularization which effectively trains node representation, so that it can alleviate the confirmation bias~\cite{confirmation} caused by inaccurate pseudo labels. Moreover, since it is severe problem which is detrimental to the performance of pseudo-labeling-based semi-superivsed methods, we additionally introduce the confidence-based label-guided consistency regularization:}
% Although $\mathbf{\tilde{p}^{'}}_i$ is computed based on the high-quality 
{However, naively minimizing the above loss would incur confirmation bias~\cite{confirmation} due to inaccurate $\mathbf{\tilde{p}^{'}}_i$ computed for unlabeled nodes (i.e., $V_U)$, which is detrimental to the performance of pseudo-labeling-based semi-supervised methods.
To this end, we introduce a confidence-based label-guided consistency regularization:}
\begin{equation}
% \small
\label{eqn:lc}
\mathcal{L}_{\text{LC}} = \frac{1}{|V_\text{conf}|}\sum_{v_i\in V_\text{conf}}H(\tilde{\mathbf{p}}_{i}^{'},\tilde{\mathbf{p}}_{i})
+ \frac{1}{|V_L|}\sum_{v_i\in V_L}H(\mathbf{Y}_i,\tilde{\mathbf{p}}_{i})
% \end{split}
\end{equation}
% \begin{equation}
% \small
% \label{eqn:lc}
% \mathcal{L}_{\text{LC}} = \frac{1}{|V_\text{conf}|}\sum_{v_i\in V_\text{conf}}H(Sharpen(\tilde{\mathbf{p}}_{i}^{'}),\tilde{\mathbf{p}}_{i})
% + \frac{1}{|V_L|}\sum_{v_i\in V_L}H(\mathbf{Y}_i,\tilde{\mathbf{p}}_{i})
% % \end{split}
% \end{equation}
where $V_\text{conf} =\{v_i|\mathbbm{1}(\max(\tilde{\mathbf{p}}_{i}^{'})>\nu) = 1, \forall v_i\in V_U\}$ is the set of nodes with confident predictions, $\nu$ is the threshold for determining whether a node has confident prediction, and $\mathbbm{1}(\cdot)$ is an indicator function.
% $\nu$ is a threshold and 
% $N_{C}$ is the number of confident nodes, i.e., 
% $V_C =\{v_i|\mathbbm{1}(\max(\tilde{\mathbf{p}}_{i}^{'})>\nu) = 1, \forall v_i\in V\}$ where $N_C=|V_C|$.
$\tilde{\mathbf{p}}_{i}^{'}$ is considered to be confident if its maximum element is greater than $\nu$. 
We argue that setting $\nu$ to a high value helps alleviate confirmation bias~\cite{confirmation, fixmatch} by enforcing only high-quality target distribution $\tilde{\mathbf{p}}_{i}^{'}$ to be able to contribute to Eqn.~\ref{eqn:lc}.
It is important to note that we apply a relatively weak augmentation for graph $\tilde{G}^{'}=(\tilde{\mathbf{A}}^{'}, \tilde{\mathbf{X}}^{'})$ that is used to compute $\tilde{\mathbf{p}}_{i}^{'}$, because aggressive augmentations (e.g., dropping more than half of edges) drastically change the semantics of a graph, which may eventually incur inaccurate $\tilde{\mathbf{p}}_{i}^{'}$.
A further benefit of the label-guided consistency regularization defined in Eqn.~\ref{eqn:lc} is that since the class distributions are computed regardless of the structural information of graphs, the label information can be effectively propagated to distant nodes, whereas existing GNNs suffer from ineffective propagation incurred by the over-smoothing issue~\cite{deeper}.
% By doing so, the label information can be effectively propagated to distant nodes. 
Moreover, to break the symmetry of the model architecture thereby preventing the collapsing of node representations to trivial solutions, we stop gradient for the target distribution (i.e., $\tilde{\mathbf{p}}^{'}$), and only update the parameters associated with the prediction distribution (i.e., $\tilde{\mathbf{p}}$).
% On the implementation level, we stop gradient for target distribution to construct asymmetry structure, equation 4. 

\smallskip
\noindent\textbf{4) Final Objective.}
Finally, we combine $\mathcal{L}_{\text{NC}}$ and $\mathcal{L}_{\text{LC}}$ with coefficients ${\lambda_{1}}$ and ${\lambda_{2}}$ to compute the final objective function as follows:
\begin{equation}
\mathcal{L}_{\text{Training}} = \lambda_{1}\mathcal{L}_{\text{NC}} + \lambda_{2}\mathcal{L}_{\text{LC}} + \mathcal{L}_{\text{sup}}
\end{equation}
We add the cross-entropy loss, i.e., $\mathcal{L}_\text{sup}$, defined over a set of labeled nodes $V_L$.
The overall pipeline of~\proposed~is shown in Fig.~\ref{fig:overall}.

% %%%%%%%%%%%%%%%%%%%%
\begin{table}[t]
\centering
% \small
\caption{Statistics for datasets used for experiments.}
\vspace{-2ex}
\renewcommand{\arraystretch}{0.95}
\begin{tabular}{c|ccccc}
\noalign{\smallskip}\noalign{\smallskip}\hline
& \# Nodes & \# Edges & \# Features & \# Classes \\
\hline
Cora & 2,708 & 5,429 & 1,433 & 7 \\
Citeseer & 3,327 & 4,732 & 3,703 & 6 \\
Pubmed & 19,717 & 44,338 & 500 & 3 \\
Am. Comp. & 13,752 & 245,861 & 767 & 10 \\
Am. Photos & 7,650 & 119,081 & 745 & 8 \\
\hline
\end{tabular}
\label{tab:dataset}
\vspace{-1ex}
\end{table}
% %%%%%%%%%%%%%%%%%%%%

\section{EXPERIMENTS}
\noindent{\textbf{Datasets.}}
To verify the effectiveness of~\proposed,
% on the real-world datasets, 
we conduct extensive experiments on five widely used datasets (Table~\ref{tab:dataset}) including three citation networks (Cora, Citeseer, Pubmed)~\cite{cora_cite_pub} and two co-purchase networks (Amazon Computers, Amazon Photos)~\cite{amazon}.
% which node types are documents, and items and edges indicate citation and frequently co-purchase relationship, respectively. 
% Detailed statistics for each dataset can be found in Table~\ref{tab:dataset}.

\smallskip
\noindent{\textbf{Baselines.}}
We compare~\proposed~with fifteen baseline methods. (i) \emph{Two conventional methods}: \textbf{MLP} and \textbf{LP}~\cite{LPA}. (ii) \emph{Five graph convolution-based methods}: \textbf{GCN}~\cite{GCN} and \textbf{GAT}~\cite{GAT}.
% are representative GNN models. 
\textbf{SGC}~\cite{SGC} simplifies GCN by removing repeated feature transformations and nonlinearities. \textbf{APPNP}~\cite{APPNP} and \textbf{GRAND}~\cite{GRAND} alleviate the limited receptive field issue of existing message passing models. (iii) \emph{Two self-supervised methods}: \textbf{GRACE}~\cite{GRACE} and \textbf{BGRL}~\cite{BGRL} maximize the agreement of the representations of the same nodes from two augmented views\footnote{For fair comparisons with~\proposed, we extend them to semi-supervised setting by adding the conventional supervised loss, i.e., cross-entropy loss.}.
% , motivated by SimCLR and BYOL~\cite{simclr, byol}, respectively. 
% We adopt GCN as the backbone of both methods and  
(iv) \emph{Three label-efficient methods}: \textbf{GLP}~\cite{GLP_IGCN} and \textbf{IGCN}~\cite{GLP_IGCN} apply a low-pass graph filter to message propagation to achieve label efficiency. \textbf{CGPN}~\cite{cgpn} leverages Graph Poisson Network to effectively spread limited labels to the whole graph, and also utilizes contrastive loss to leverage information on unlabeled nodes. (v) \emph{Three pseudo-labeling-based methods}: \textbf{Co-training}~\cite{deeper}, \textbf{Self-training}~\cite{deeper} 
% aim to create confident pseudo-labels, 
and \textbf{M3S}~\cite{M3S}.
% extends self-training by filtering out pseudo-labels that do not align with the clustering assignments. 

\smallskip
\noindent{\textbf{Evaluation Protocol.}}
We randomly create 20 splits to evaluate the effectiveness of~\proposed~on practical few-label setting. For citation networks, we closely follow the evaluation protocol of~\cite{deeper, investigating}.
% train nodes are masked upon each label rate correspondingly following prior work ~\cite{deeper}, 
For co-purchase networks, we evaluate models with \{0.15\%, 0.2\%, 0.25\%\} training label rates
to match with the average number of labeled nodes per class of the citation networks.
% train nodes are masked upon \{0.15\%, 0.2\%, 0.25\%\} label rate 
% to set the average number of labeled nodes per class to 1\textasciitilde4. 
The remaining nodes are split into 1:9, and each split is used for validation and test, respectively. We report the averaged test accuracy when the validation accuracy is the highest.
For baseline methods including~\proposed, we search hidden dimension, learning rate, weight decay, and dropout ratio in \{32, 64, 128\}, \{0.1, 0.01, 0.05, 0.001, 0.005\}, \{1e-2, 1e-3, 1e-4, 5e-4\}, and \{0, 0.3, 0.5, 0.8\}, respectively, while other hyperparameter configurations are taken from each work. 
Additionally, we conduct an extensive grid search for unreported hyperparameters for fair comparisons.
% ~\proposed~also shares same search space for the aforementioned hyperparameters. 
As mentioned in Sec.~\ref{sec:method}, for~\proposed, we apply relatively weaker augmentations on $\tilde{G}^{'}$ compared with $\tilde{G}$, and thus we search the augmentation hyperparameters of node feature masking and partial edge dropping in \{0.2, 0.3, 0.4\} for $\tilde{G}^{'}$, and \{0.5, 0.6\} for $\tilde{G}$. {Moreover, the temperature $\tau$ is fixed to 0.1}, the threshold $\nu$ is searched in \{0.8, 0.9\}, and the balance hyperparameters, i.e., $\lambda_{1}$ and $\lambda_{2}$, are searched in \{0.5, 1.0, 2.0\}.

\begin{table}[t]
	\centering
% 	\small
	\renewcommand{\arraystretch}{0.99}
	\caption{Performance on similarity search. (Sim@$K$: Average ratio among $K$-NNs sharing the same label as the query node.)}
	\vspace{-1ex}
	\begin{tabular}{p{1.3cm}|p{1.0cm}|ccc}
		\multicolumn{2}{c|}{} & GRACE & BGRL &~\proposed~\\ \hline \hline
		\multirow{2}{*}{Cora} & Sim@5 & 0.8146 & 0.8047 & \textbf{0.8222} \\
		& Sim@10 & 0.7947 & 0.7823 & \textbf{0.7984} \\ \hline
		\multirow{2}{*}{Citeseer} & Sim@5 & 0.6407 & 0.6623 & \textbf{0.6810} \\
		& Sim@10 & 0.6147 & 0.6396 & \textbf{0.6621} \\ \hline
		\multirow{2}{*}{Pubmed} & Sim@5 & 0.7571 & \textbf{0.7815} & 0.7110 \\
		& Sim@10 & 0.7493 & \textbf{0.7733} & 0.7015 \\ \hline
		\multirow{2}{*}{Am.Comp} & Sim@5 & 0.8091 & 0.8335 & \textbf{0.8351} \\
		& Sim@10 & 0.7965 & 0.8211 & \textbf{0.8246} \\ \hline
		\multirow{2}{*}{Am.Photos} & Sim@5 & 0.8831 & 0.8886 & \textbf{0.9004} \\
		& Sim@10 & 0.8761 & 0.8881 & \textbf{0.8937} \\ \hline
	\end{tabular}
	\vspace{-3ex}
	\label{tab:similarity_search}
\end{table}

\subsection{Performance Analysis}
Table ~\ref{tab:main_table} shows the performance of various methods in terms of node classification over various label rates. We have the following observations: 
\textbf{1)}~\proposed~outperforms methods that use advanced GNN encoders (i.e., APPNP and GRAND), and label efficient GNNs that effectively propagate label information to distant nodes (e.g., GLP and IGCN) and leverage unlabeled nodes with contrastive loss (i.e., CGPN). 
We argue that the label-guided consistency regularization helps ~\proposed~ to effectively propagate the label information to distant nodes, thereby learning class discriminative representations. Note that~\proposed~especially outperforms other methods when the label rate is small demonstrating the robustness of~\proposed~(also refer to Fig.~\ref{fig:teaser}).
% for low-degree nodes which suffer from the insufficient supervision~\cite{investigating}.
% as label-guided consistency regularization help to effectively propagate label information to the distant nodes, \proposed~ also can learn discriminative representation for low-degree nodes which suffer from the insufficient supervision~\cite{investigating}. 
% \textbf{1)} Even if ~\proposed~ adopt GCN as backbone encoder, it outperforms advanced encoder (e.g.APPNP and GRAND), and CGPN which effectively propagates label information to distant nodes and leveraging unlabed nodes with contrastive loss. We argue that as label guided consistency regularization help to effectively propagate label information to the distant nodes, \proposed~ also can learn discriminative representation for low-degree nodes which suffer from the insufficient supervision~\cite{investigating}. 
\textbf{2)} Even though~\proposed~adopts a simple self-supervised loss with a single shared GNN-based encoder, i.e. node-wise consistency regularization,~\proposed~outperforms advanced self-supervised methods, i.e. GRACE and BGRL. We attribute this to the label-guided consistency loss that groups nodes that belong to the same class by leveraging the given few labeled nodes.
% because \proposed~ groups nodes belong to same class by leveraging given label information through label guided consistency loss. 
\textbf{3)} To empirically verify the benefit of the label-consistency regularization of~\proposed, we compare the similarity search results of GRACE, BGRL and~\proposed~in Table ~\ref{tab:similarity_search}. Since GRACE and BGRL consider node labels through conventional supervised loss with advanced self-supervised loss, whereas~\proposed~does so through the label-guided consistency regularization with a simple self-supervised loss, the superior performance of~\proposed~implies the benefit of the label-guided consistency regularization of~\proposed~despite its simple self-supervised loss. We indeed observe that~\proposed~generally outperforms GRACE and BGRL, which corroborates the benefit of the label-consistency regularization for learning class discriminative node representations.
% \textbf{2)} Even though, we use much simple self-supervised loss i.e. Node-wise Consistency Regularization, \proposed~ outperforms self-supervised based counterparts ,i.e. GRACE, BGRL, because \proposed~ groups nodes belong to same class by leveraging given label information through label guided consistency loss. To verify this argument, we additionally compare similarity search performance with above methods in Table ~\ref{tab:similarity_search}. It can show that \proposed~ learn much discriminative representation which nodes belong to same class group together. 
\textbf{4)} We observe that the pseudo-labeling-based methods, i.e., Co-training, Self-training and M3S, generally outperform vanilla GCN, which solely relies on the given label information. On the other hand,~\proposed~outperforms these methods without relying on the pseudo-labeling techniques. This implies that the pseudo-labeling techniques should be adopted with particular care as they may introduce incorrect labels.
% Although the pseudo-labeling methods generally outperform vanilla GCN,~\proposed~outperforms these methods by using only given label information because they have inherent limitation originated from incorrect pseudo-labels.
\textbf{5)}~\proposed~shows relatively low performance on Pubmed dataset, which contains a small number of classes, i.e., 3 classes.
Since~\proposed~assigns class distributions based on the similarity with labeled nodes in different classes, having more classes leads to more class discriminative node representations.
% Pubmed contains only a small number of classes i.e. 3 classes, which incur the lack of information when assigning class distribution based on similarity with labeled nodes in different class.
\begin{figure}[t]
  \includegraphics[width=0.99\linewidth]{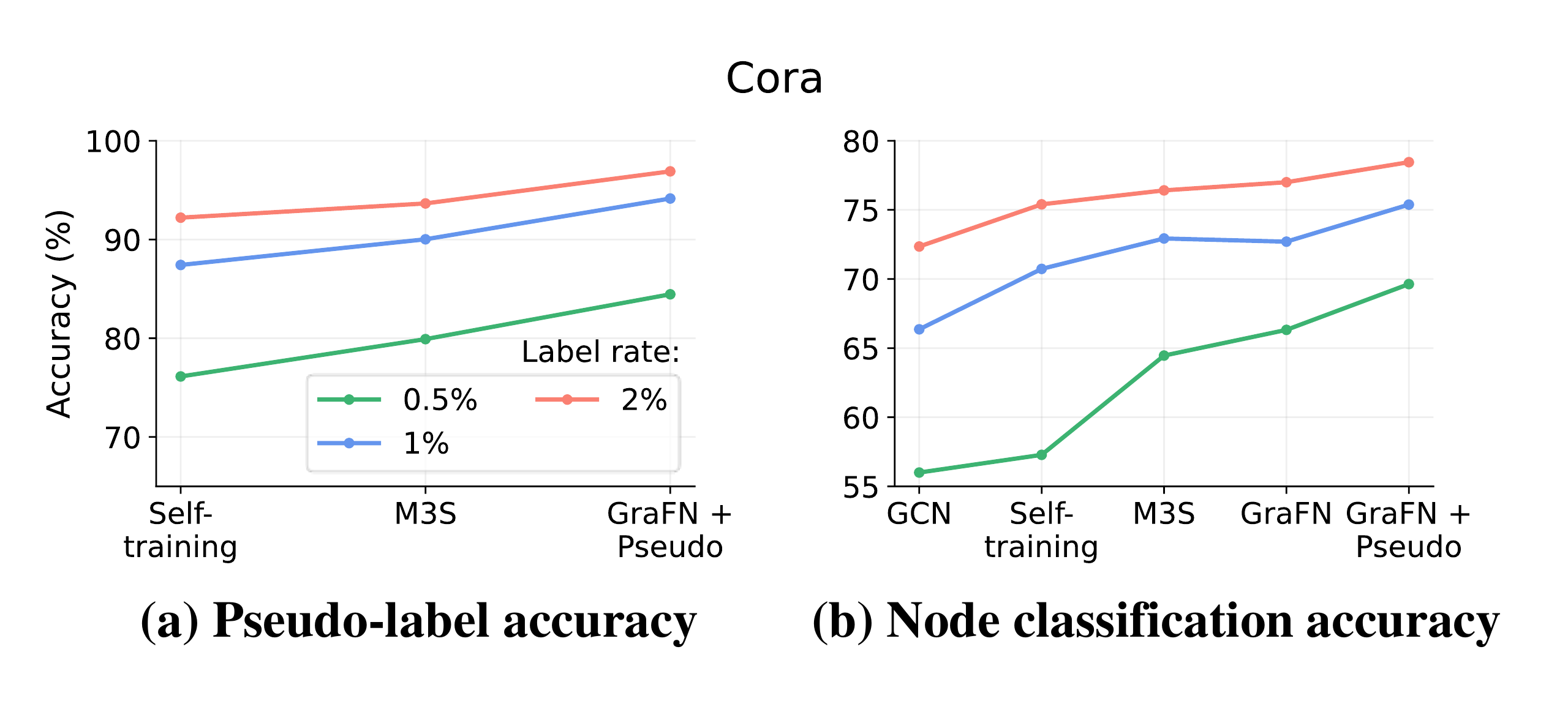}
  \vspace{-3ex}
  \caption{Accuracy of pseudo-labeling and node classification.}
    \label{fig:pseudo}
\vspace{-3ex}
\end{figure}

\smallskip
\noindent{\textbf{Adopting Pseudo-Labeling to~\proposed.}}
Since~\proposed~learns class discriminative node representations with few labels, we hypothesized that the confirmation bias~\cite{confirmation} of the pseudo-labeling technique suffered by existing methods would be alleviated when the pseudo-labeling technique is adopted to~\proposed.
Fig.~\ref{fig:pseudo}(a) indeed demonstrates that adopting the pseudo-labeling technique to~\proposed~gives the best pseudo-labeling accuracy, which in turn results in further improvements of~\proposed~in terms of node classification (Fig.~\ref{fig:pseudo}(b)).

\begin{figure}[t]
  \includegraphics[width=0.99\linewidth]{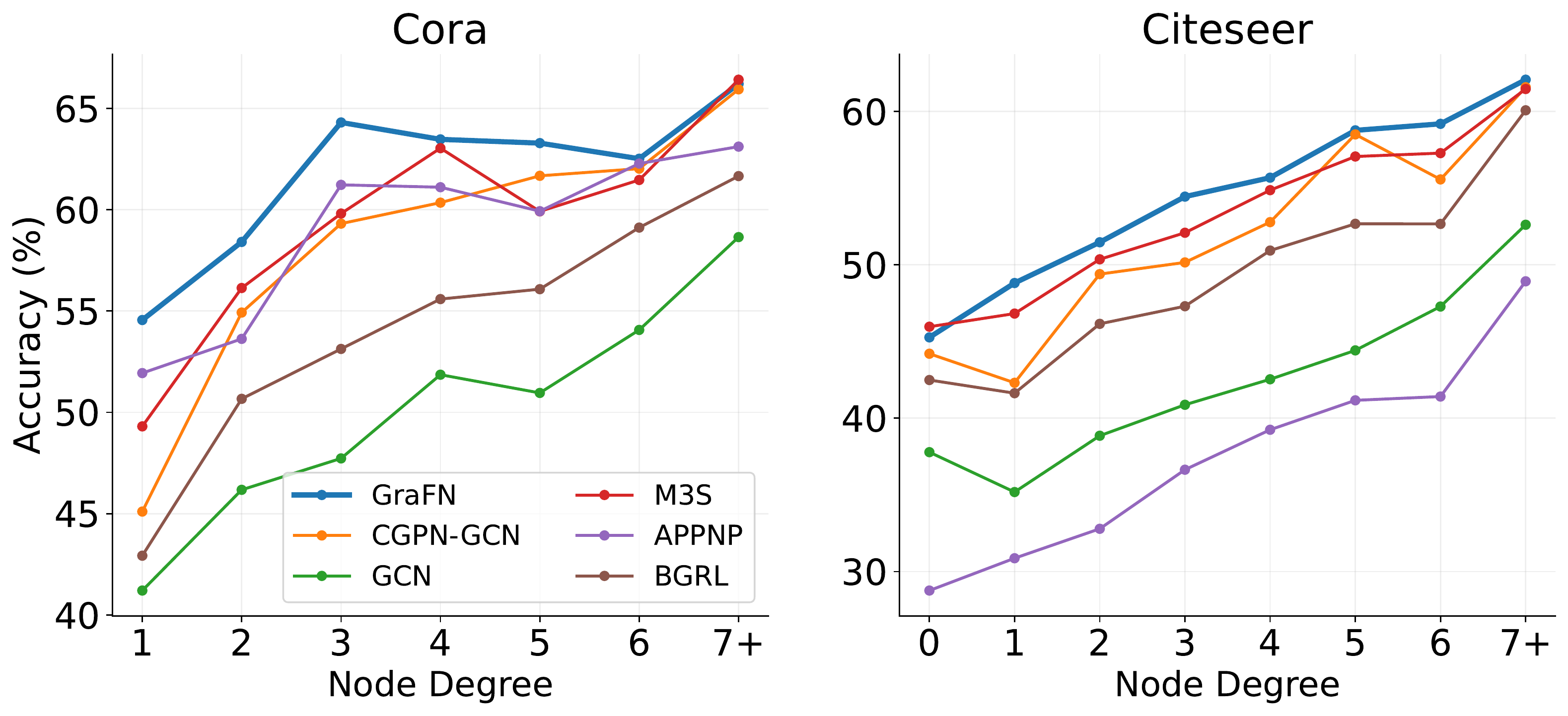}
  \vspace{-1ex}
  \caption{Node classification results on various node degrees.}
  \label{fig:Degree}
\vspace{-1ex}
\end{figure}

\smallskip
\noindent{\textbf{Performance Comparison on Different Node Degrees. }}
In most real-world graphs, the node degrees follow a power-law distribution, which means that the majority of nodes are of low-degree. 
Since the training of GNNs is based on the neighborhood aggregation scheme, high-degree nodes receive more information than low-degree nodes, which eventually leads to the models overfit to high-degree nodes and underfit to low-degree nodes~\cite{investigating}. 
% This is especially problematic for training GNN-based methods that work by aggregating neighbors 
% since their parameters are learned by minimizing the cross-entropy on the labeled nodes, and thus high-degree nodes receive more supervision than low-degree nodes incurring overfitting to high-degree nodes and underfitting to low-degree nodes~\cite{investigating}. 
Fig.~\ref{fig:Degree} demonstrates that the node classification accuracy is indeed highly correlated with the node degree, i.e., high-degree nodes tend to result in better classification performance.
% a node with low degrees is less likely to be linked to labeled nodes than a high one. This tendency brings out a performance degeneracy on low degree nodes in many existing GCN-based methods since they learn their parameters by minimizing a classification loss (e.g. Cross-Entropy) on labeled nodes, thereby nodes connected to labeled ones can directly receive supervision information by back-propagation \cite{investigating}. In 
% Figure ~\ref{fig:Degree}, we experimentally demonstrate the tendency, there is a high correlation between classification accuracy of nodes and their degree. 
Furthermore, we observe that~\proposed~greatly outperforms other methods for low-degree nodes, while showing a comparable performance for high-degree nodes (i.e. degree $\geq$  7). 
We attribute the superior performance of~\proposed~on low-degree nodes to the label-guided consistency regularization that 
% We argue that the label-guided consistency regularization of~\proposed~
promotes the supervision information to be evenly spread over the unlabeled nodes regardless of their node degree.

\begin{figure}[h]
\vspace{-1ex}
  \includegraphics[width=0.99\linewidth]{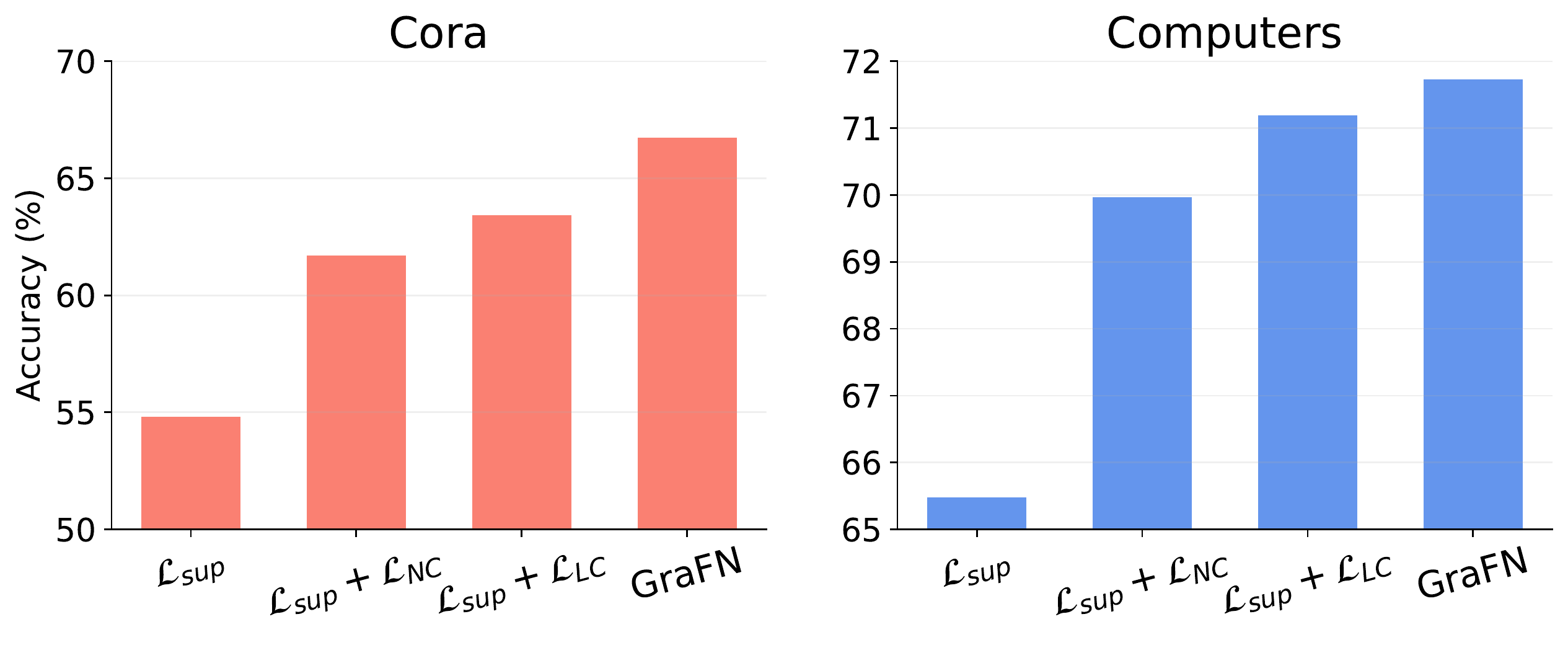}
  \vspace{-1ex}
  \caption{Ablation study on~\proposed.}
  \vspace{-3ex}
  \label{fig:ablation}
\end{figure}

\subsection{Ablation Studies}
To evaluate each component of \proposed, we conduct ablation studies on Cora and Computers datasets on the lowest label rate, i.e. 0.5\%, 0.15\%, respectively. In Fig~\ref{fig:ablation}, we have the following observations: 
\textbf{1)} Only using the supervised cross-entropy loss shows poor performance because it suffers from overfitting and ineffective propagation of supervisory signal. 
\textbf{2)} Using both node-wise and label-guided consistency regularization (i.e.,~\proposed) is more beneficial than using either one of them. We argue that this is because these two losses are complementary. More precisely, using only the node-wise loss cannot fully leverage the given label information, whereas using only the label-guided regularization loss can suffer from inaccurate target distribution, which incurs confirmation bias.

\section{CONCLUSIONS}
In this paper, we present a novel semi-supervised method for graphs that learns class discriminative node representations with only few labeled nodes.~\proposed~ not only exploits the self-supervised loss, i.e. node-wise consistency regularization, but also ensures nodes that belong to the same class to be grouped together by enforcing unlabeled nodes to be consistently close with a certain class of labeled nodes on differently augmented graphs. Through extensive experiments on real-world graphs, we show that~\proposed~outperforms existing state-of-the art methods given few labeled nodes. Moreover, it is worth noting that~\proposed~1) alleviates underfitting problem of low-degree nodes by propagating label information to distant nodes, and 2) enjoys further improvements by adopting the pseudo-labeling technique.
% In this paper, we propose a novel self-supervised learning-based method to learn class discriminative node representation with only few labeled nodes. To do this, \proposed~ not only exploit self-supervised loss, i.e. node-wise consistency regularization but also ensuring nodes belong to same class grouped together by enforcing unlabeled nodes to be consistently close with certain class of labeled nodes on differently augmented graph. Through extensive experiments on real world datasets, we show that \proposed~ outperforms existing state-of-the art methods given few label information. Moreover, it is worth noting that 1) \proposed~ alleviates underfitting problem for low degree nodes by propagating label information to distant nodes and 2) can enjoy further improvement by adopting pseudo-labeling technique.

\smallskip
\noindent\textbf{Acknowledgements.}
This work was supported by the NRF grant funded by
the MSIT (No.2021R1C1C1009081), and the IITP grant
funded by the MSIT (No.2019-0-00075, Artificial Intelligence Graduate School Program (KAIST)).

% \newpage
%%
%% The next two lines define the bibliography style to be used, and
%% the bibliography file.

%\bibliographystyle{ACM-Reference-Format}
%\bibliography{reference}

\bibliographystyle{ACM-Reference-Format}
\balance
\bibliography{reference}

%%
%% If your work has an appendix, this is the place to put it.
%\appendix

%\section{Research Methods}

\end{document}